\pdfoutput=1

\documentclass[11pt]{article}

\usepackage[preprint]{acl}

\usepackage{times}
\usepackage{latexsym}

\usepackage[T1]{fontenc}

\usepackage[utf8]{inputenc}
\usepackage{microtype}
\usepackage{inconsolata}
\usepackage{graphicx}

\usepackage{caption}    
\usepackage{multirow}
\usepackage{tabularx}
\usepackage{booktabs}
\definecolor{citecolor}{HTML}{0071BC}
\definecolor{linkcolor}{HTML}{ED1C24}
\definecolor{LGray}{gray}{0.97}
\usepackage{multicol}
\usepackage{colortbl}
\usepackage{xcolor}
\definecolor{tabhighlight}{HTML}{e5e5e5}
\usepackage{color}
\usepackage{xspace}
\usepackage{pifont}
\usepackage{colortbl} 
\usepackage{subcaption}

\usepackage{tcolorbox}
\tcbuselibrary{listings}
\usepackage{listings}
\usepackage{makecell}
\usepackage{parskip}

\hypersetup{urlcolor=blue} 

\usepackage{pgfplots}
\usepackage{caption}
\pgfplotsset{compat=1.18}

\usepackage{fix-cm}

\definecolor{darkgreen}{rgb}{0.0, 0.5, 0.0}  
\newcommand{\cmark}{\textcolor{darkgreen}{\scalebox{1}[1.0]{\ding{51}}}}
\newcommand{\xmark}{\textcolor{red}{\ding{55}}}  

\definecolor{darkblue}{RGB}{0,0,120}  

\hypersetup{
  colorlinks=true,
  citecolor=darkblue,
  linkcolor=darkblue,
  urlcolor=darkblue
}

\title{
    \begin{minipage}{0.12\textwidth} 
        \raggedleft
        \includegraphics[height=1.75cm]{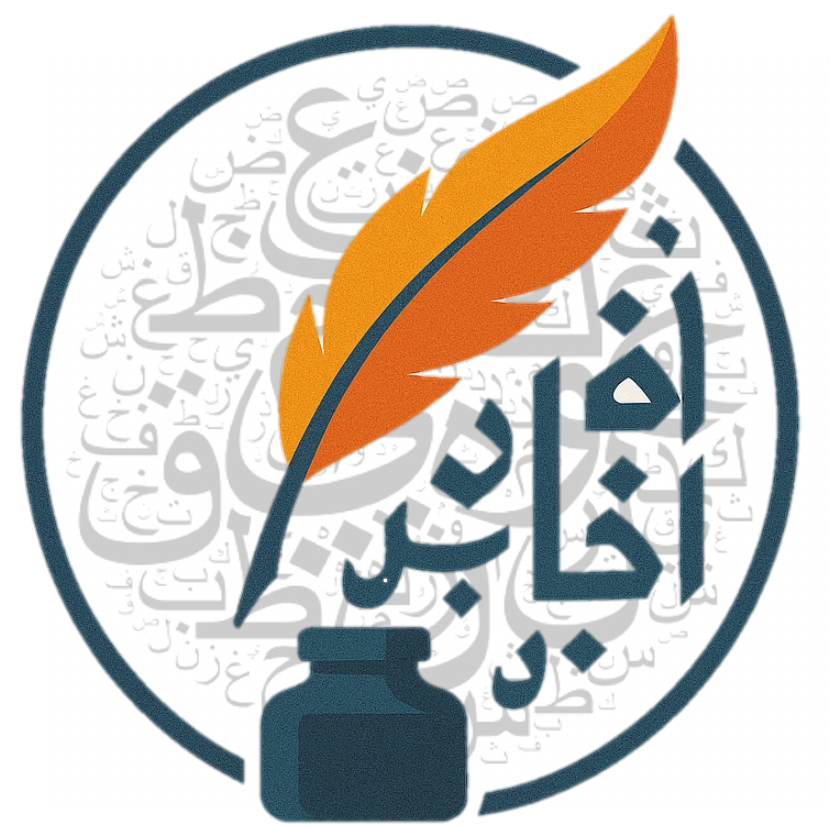} 
    \end{minipage}%
    \hspace{0.005cm} 
    \begin{minipage}{0.8\textwidth} 
        \centering
        \textbf{DuwatBench: Bridging Language and Visual Heritage through an Arabic Calligraphy Benchmark for Multimodal Understanding}
    \end{minipage}
}

\author{\\ {Shubham Patle}\textsuperscript{1$\dagger$} \quad{Sara Ghaboura}\textsuperscript{1$\dagger$} \quad{Hania Tariq}\textsuperscript{2}\quad {Mohammad Usman Khan}\textsuperscript{3}\\ \quad{Omkar Thawakar}\textsuperscript{1} \quad
     {Rao Muhammad Anwer}\textsuperscript{1}\quad
     {Salman Khan}\textsuperscript{1,4}\\
     \fontsize{11pt}{12pt}\selectfont \textsuperscript{1}Mohamed bin Zayed University of AI, \textsuperscript{2}NUCES,
     \textsuperscript{3}NUST, \textsuperscript{4}Australian National University\\
     \fontsize{10pt}{12pt}\selectfont \{{shubham.patle, sara.ghaboura, omkar.thawakar}\}@mbzuai.ac.ae \\
 {\hypersetup{urlcolor=blue}
\fontsize{11pt}{12pt}\selectfont \href{https://mbzuai-oryx.github.io/DuwatBench/}{https://mbzuai-oryx.github.io/DuwatBench/}} \\ }

\begin{document}
\maketitle
\begingroup
\renewcommand{\thefootnote}{\fnsymbol{footnote}}
\footnotetext[2]{Equal contribution.}
\endgroup

\begin{abstract}
Arabic calligraphy represents one of the richest visual traditions of the Arabic language, blending linguistic meaning with artistic form. Although multimodal models have advanced across languages, their ability to process Arabic script, especially in artistic and stylized calligraphic forms, remains largely unexplored. To address this gap, we present DuwatBench, a benchmark of 1,272 curated samples containing about 1,475 unique words across 6 classical and modern calligraphic styles, each paired with sentence-level detection annotations. The dataset reflects real-world challenges in Arabic writing, such as complex stroke patterns, dense ligatures, and stylistic variations that often challenge standard text recognition systems.
Using DuwatBench, we evaluated 13 leading Arabic and multilingual multimodal models and showed that while they perform well in clean text, they struggle with calligraphic variation, artistic distortions, and precise visual–text alignment. By publicly releasing DuwatBench and its annotations, we aim to advance culturally grounded multimodal research, foster fair inclusion of Arabic language and visual heritage in AI systems, and support continued progress in this area. 
Our dataset \footnote{\url{https://github.com/mbzuai-oryx/DuwatBench}} and code\footnote{\url{https://huggingface.co/datasets/MBZUAI/DuwatBench}} are publicly available.
\end{abstract}

\begin{figure}[t]
\centering
\includegraphics[width=0.98\linewidth,height=5.5cm]{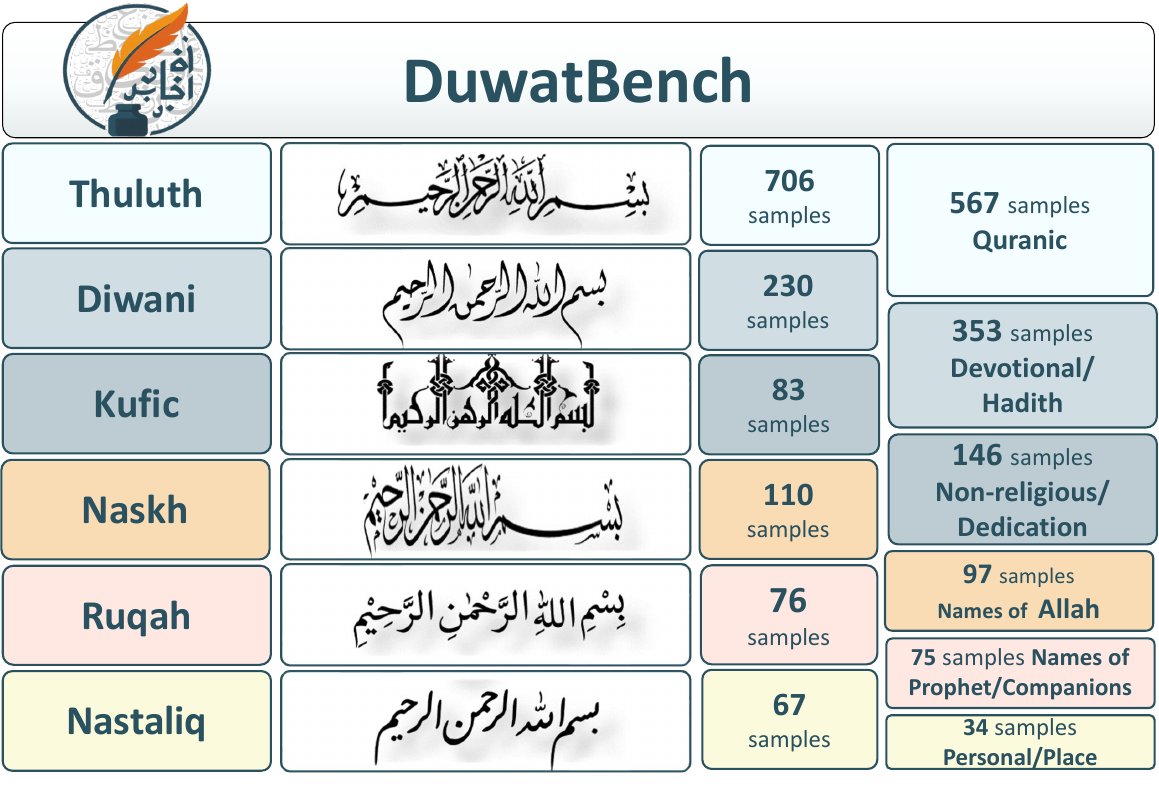}
\vspace{-0.75em}
\caption{
\small
\textbf{DuwatBench Taxonomy and Distribution.} DuwatBench encompasses six principal Arabic calligraphy styles (Thuluth, Diwani, Kufic, Naskh, Ruq'ah, and Nasta'liq). The benchmark includes a distribution of collected samples ranging from 706 in Thuluth to 67 in Nasta'liq. Beyond style diversity, the taxonomy incorporates categories, including non-religious terms, Quranic words, devotional expressions and hadith, names of the Prophet and companions, names of Allah, and person or place names. This organization provides a structured basis for evaluating both the visual variation and semantic depth of Arabic calligraphy.}
\vspace{-1.25em}
\label{fig:taxonomy}
\end{figure}

\begin{table*}[!ht]
\centering
\renewcommand{\arraystretch}{1.2}
\setlength{\tabcolsep}{6pt}
\resizebox{\textwidth}{!}{
\begin{tabular}{l|cccccccc}
\toprule
\textbf{Feature} & \textbf{Allaf} & \textbf{Adam} & \textbf{Sal\&King} & \textbf{Kaoudja} & \textbf{Calliar} & \textbf{MOJ-DB} & \textbf{HICMA} & \textbf{DuwatBenach \small\emph{(Ours)}} \\
\hline
Real-world images & \cmark & \cmark & \cmark & \cmark & \xmark & \cmark & \cmark & \cmark \\
Style diversity (multiple styles) & \xmark & \cmark & \cmark & \cmark & \cmark & \xmark & \cmark & \cmark \\
Word / Sentence-level text & \cmark & \xmark & \xmark & \cmark & \cmark & \cmark & \cmark & \cmark \\
Full text transcriptions & \xmark & \xmark & \xmark & \xmark & \cmark & \cmark & \cmark & \cmark \\
Bounding boxes for detection & \xmark & \xmark & \xmark & \xmark & \xmark & \xmark & \xmark & \cmark \\
Complex artistic backgrounds & \cmark & \xmark & \xmark & \xmark & \xmark & \cmark & \cmark & \cmark \\
Publicly available & \xmark & \xmark & \xmark & \xmark & \cmark & \cmark & \cmark & \cmark \\
\bottomrule
\end{tabular}
}
\vspace{-0.5em}
\caption{
\small
\textbf{Comparison of existing Arabic calligraphy datasets.} Existing resources typically focus on isolated aspects such as limited style coverage, word- or sentence-level annotations, or constrained availability. In contrast, DuwatBench uniquely integrates multiple underrepresented dimensions like script diversity, word- and sentence-level transcriptions, bounding boxes for detection, and complex artistic backgrounds, while ensuring public accessibility. \small Allaf \cite{allaf2016automatic}, Adam \cite{adam2017based}, Sal\&King \cite{salamah2018towards}, Kaoudja \cite{kaoudja2022arabic}, Calliar \cite{alyafeai2022calliar}, MOJ-DB\cite{zoizou2022moj}, HICMA \cite{ismail2023hicma}.}
\label{tab:comparison_all}
\vspace{-1em}
\end{table*}

\section{Introduction}

Arabic calligraphy sits at the meeting point of language and visual art. Letters are bound, stretched, and ornamented; diacritics and ligatures introduce significant variation in glyph shapes and baselines; style rules vary across schools such as Thuluth, Diwani, Naskh, and Kufic. These properties make calligraphy culturally and technically challenging for machine perception. Prior works (\cite{lorigo2006offline}, \cite{al2017arabic}) have noted the added difficulty of Arabic script due to positional letter forms, dense ligaturing, and heavy use of diacritics, especially when written artistically rather than as plain text.

Despite growing interest in Arabic handwriting and manuscripts, existing datasets only partially address calligraphic use cases. Calliar~\cite{alyafeai2022calliar} provides pen-trace data across multiple text levels but lacks the textures and visual complexity of real artworks. HICMA~\cite{ismail2023hicma} includes real-world calligraphy images with sentence-level and style labels, but is limited to five scripts, focuses mainly on recognition, and lacks layout annotations as well as evaluation on large multimodal models (LMMs). Earlier works on style classification~\cite{kaoudja2022arabic} or letter-level corpora~\cite{adam2017based,salamah2018towards} remain too narrow to support comprehensive multimodal assessment.

Meanwhile, Arabic-capable multimodal models are advancing rapidly. Cross-lingual CLIP variants~\cite{chen2023mclip} and multilingual VLMs~\cite{geigle2023mblip} demonstrate strong retrieval and captioning capabilities across languages; Arabic-specific CLIP adaptations such as AraCLIP~\cite{al2024araclip} report clear gains on Arabic image–text benchmarks; and recent Arabic OCR and VLM systems~\cite{bhatia2024qalam, wasfy2025qari, heakl2025ainarabicinclusivelarge} show improved handling of diacritics and high-resolution text. However, none of these models have been evaluated on stylized Arabic calligraphy with complex artistic backgrounds and diverse style rules. A benchmark centered on calligraphy can thus reveal whether these systems can generalize to culturally grounded, visually demanding Arabic text, or still rely on surface-level cues.

We introduce DuwatBench, a 1.27K-sample benchmark for Arabic calligraphy designed to evaluate LMMs under realistic conditions. The dataset encompasses over 1,475 unique words drawn from both religious and cultural domains, including Quranic verses, devotional phrases, greetings, and poetic or proverbial expressions. It features 6 major calligraphic styles: Thuluth, Diwani, Naskh, Kufic, Nasta'liq, and the more modern Ruq'ah, capturing the visual diversity of Arabic writing across historical and contemporary contexts (see Figure~\ref{fig:taxonomy}). Each image is paired with transcriptions and detection-level annotations, enabling both recognition and localization analysis.

Unlike existing datasets that primarily target OCR-like tasks (Table~\ref{tab:comparison_all}), DuwatBench integrates real-world calligraphy with complex backgrounds, multiple scripts, full transcriptions, and bounding box annotations. Arabic calligraphy lies at the edge of visual–textual alignment: while meaning is preserved, artistic strokes, ligatures, and curved layouts often distort the patterns leveraged by models trained on clean Latin text. Evaluating multimodal models on DuwatBench therefore provides a rigorous test of whether Arabic or multilingual pretraining transfers to artistic calligraphy, distinguishing genuine understanding from pattern memorization.

Our contributions are threefold: (1) we release a curated benchmark of 1.27K samples spanning 6 calligraphic styles and 1,475 unique words, enriched with transcriptions and artistic contexts; (2) we introduce detection-level annotations for word localization, complementing prior classification- and OCR-based datasets; and (3) we establish systematic baselines on state-of-the-art Arabic and multilingual LMMs, revealing consistent failure cases such as style misinterpretation, sensitivity to diacritics and hamza, curved text alignment, and interference from background clutter.

DuwatBench bridges Arabic visual heritage and modern AI, enabling inclusive evaluation, style-aware modeling, and applications in preservation, education, and cultural search.

\begin{figure*}[!ht]
\centering
\includegraphics[width=\linewidth, height=11.5cm]{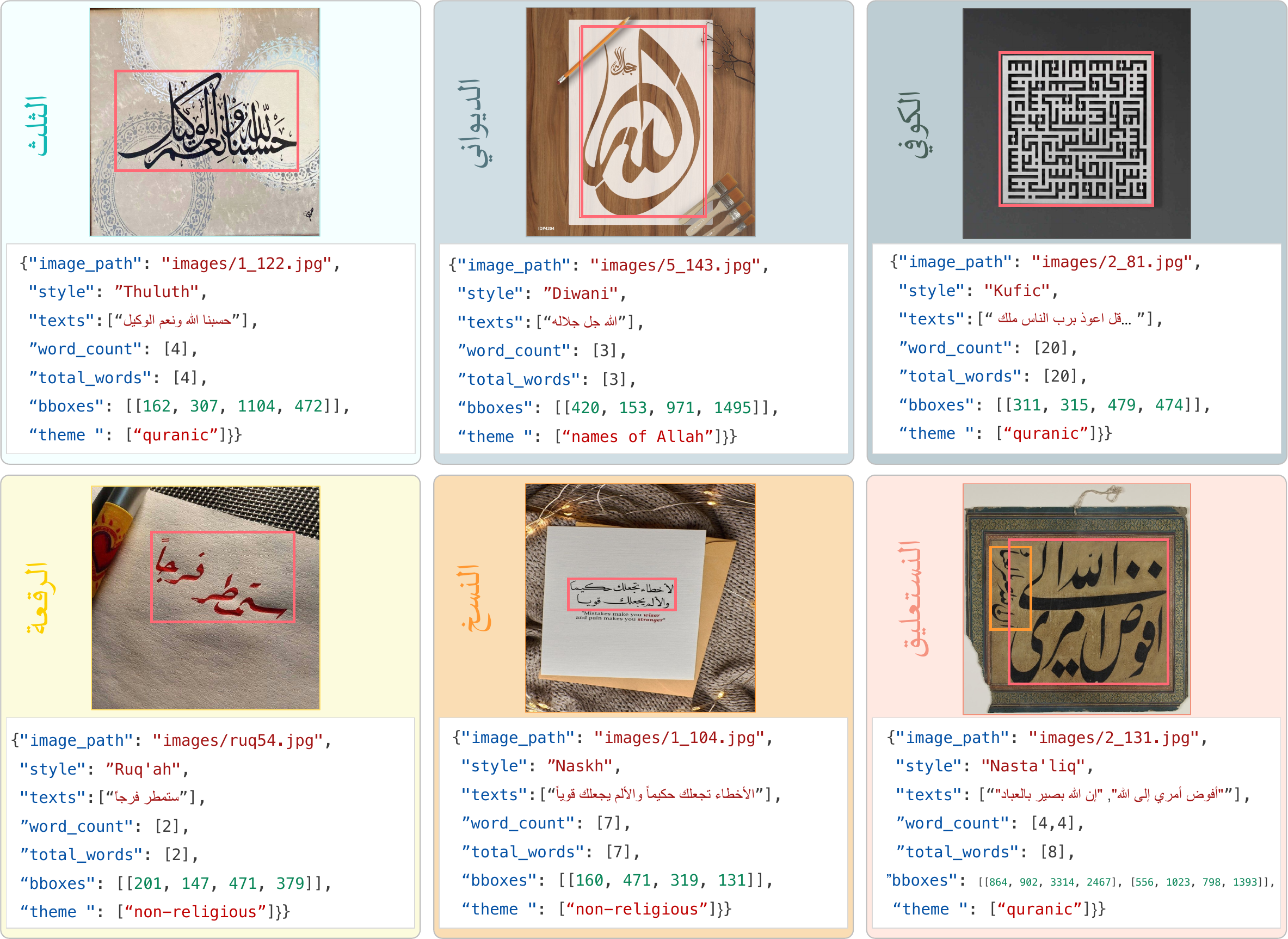}
\vspace{-1.75em}
\caption{
\small
\textbf{Examples of DuwatBench Calligraphic Styles.} Representative samples from the six calligraphic styles in DuwatBench: Thuluth, Diwani, Kufic, Ruq'ah, Naskh, and Nasta'liq. Each entry displays the artwork with bounding box annotations, transcription, and metadata such as style, text content, theme, and word count. These examples highlight the dataset’s diversity in structure, composition, and artistic context, spanning both religious and non-religious inscriptions.}
\vspace{-0.75em}
\label{fig:samples_style}
\end{figure*}

\section{DuwatBench: The Arabic Calligraphy Dataset}
\subsection{Dataset Taxonomy}

Our dataset corpus is structured along two main dimensions: textual categories and calligraphic styles (see Figure~\ref{fig:taxonomy}). On the content side, the largest portion consists of religious material, including Quranic verses, devotional invocations, and the Names of Allah, complemented by references to the Prophets and companions, as well as smaller portions of hadith and personal dedications. Non-religious expressions are also well represented, such as greetings, motifs, and cultural quotes common in decorative and public settings (see Figure~\ref{fig:theme_stat}). This balance ensures fair representation of both spiritual and cultural expressions in the dataset.

In terms of visual representation, the dataset covers 6 major calligraphic styles (Thulth, Diwani, Kufic, Nash, Ruq'ah, and Nasta'liq) with varying frequencies (see Figure~\ref{fig:style_stat}). Further detailed statistical breakdown is provided in Appendix~\ref{app:data_stat}.

\subsection{Data Collection and Candidate Selection}
\begin{figure*}[t!]
\centering
\includegraphics[width=\textwidth,height=6cm]{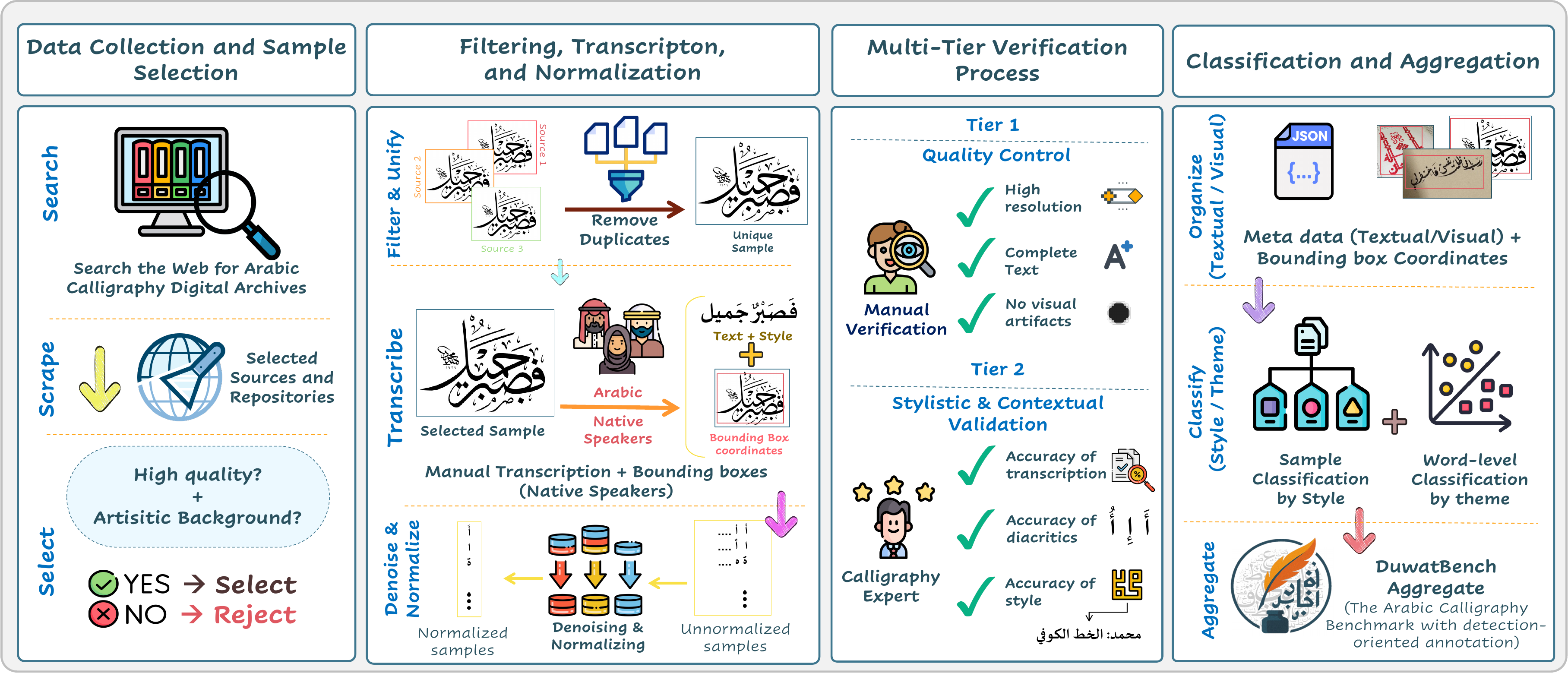}
\vspace{-1.75em}
\caption{
\small
\textbf{Data collection and verification pipeline for DuwatBench.} The process begins with sourcing candidate images from digital archives and community repositories, followed by quality-based selection and manual transcription by native Arabic speakers. Each sample is denoised, normalized, and assigned bounding boxes with text–style and theme annotations. A multi-tier validation stage involving quality control, stylistic checks, and expert review ensures both textual accuracy and calligraphic fidelity. Final samples are classified and aggregated by style and thematic category, forming the curated DuwatBench corpus.}
\label{fig:data_pipeline}
\vspace{-0.75em}
\end{figure*}

As shown in Figure~\ref{fig:data_pipeline}, the pipeline starts by sourcing candidate images from digital archives \footnote{\url{https://www.loc.gov/collections/}}\textsuperscript{,}\footnote{\url{https://digitalcollections.nypl.org/}} 
and community repositories 
\footnote{\url{https://calligraphyqalam.com/}}\textsuperscript{,}\footnote{ \url{https://freeislamiccalligraphy.com/}}\textsuperscript{,}\footnote{ \url{https://www.pinterest.com/}}. These candidates are screened for resolution, completeness, and the presence of authentic artistic backgrounds. Low-quality, blurred, or incomplete inscriptions are discarded at this stage, ensuring that only visually reliable samples move forward. We began with more than 2,950 samples, and after initial filtering and duplicate removal, the dataset was reduced to 1,285 high-quality instances, representing an overall reduction of approximately 57\%.

\subsection{Filtering, Transcription, \& Normalization}
\begin{figure}[!ht]
\centering
\includegraphics[width=0.5\textwidth,height=5.25cm]{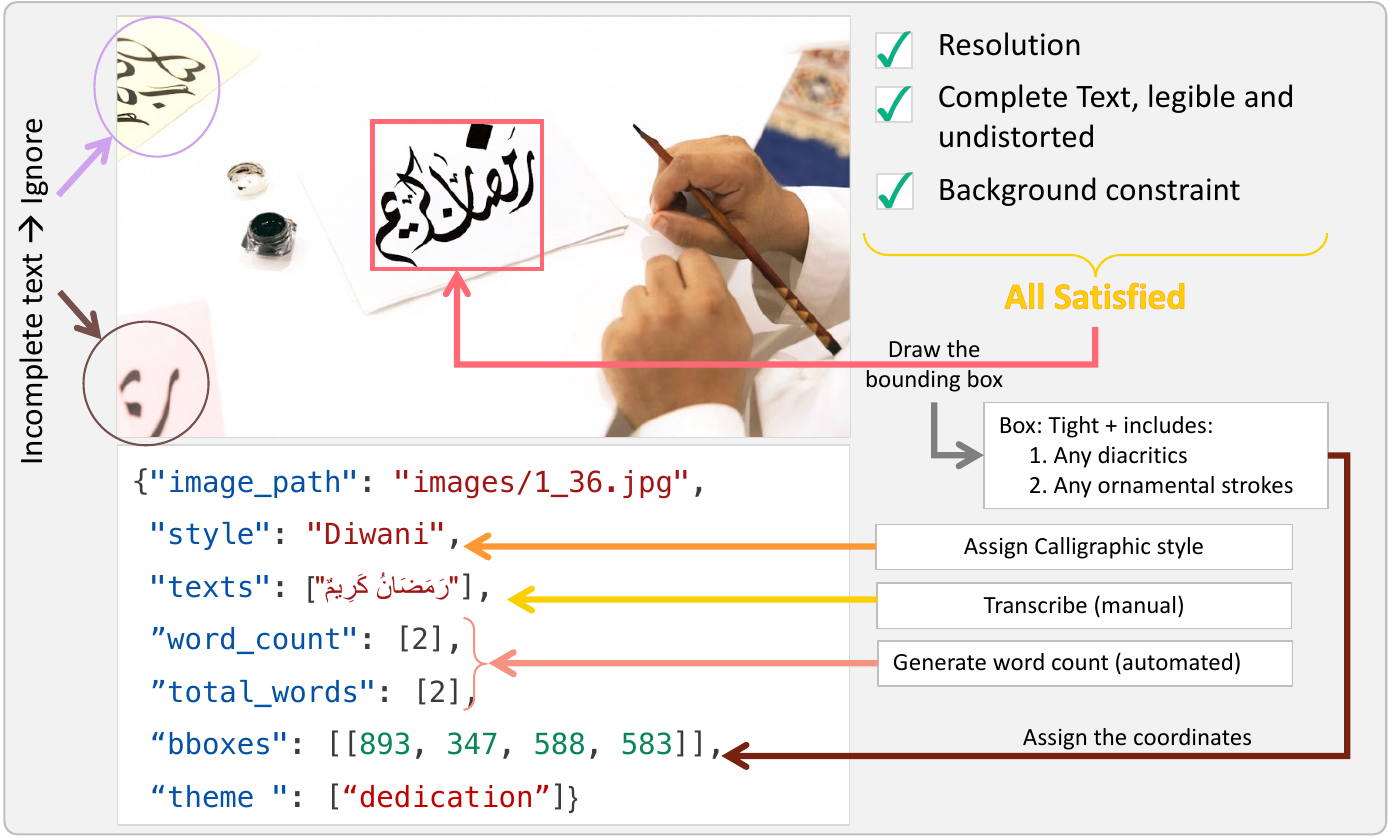}
\vspace{-1.25em}
\caption{
\small
\textbf{Annotation Workflow in DuwatBench.}
The figure illustrates the end-to-end annotation process followed in DuwatBench, from visual quality screening and bounding box drawing to transcription, style labeling, and word-count generation. Each stage ensures completeness, legibility, and consistent metadata alignment between text and visual form, supporting accurate multimodal evaluation.}
\label{fig:annotation_process}
 \vspace{-0.75em}
\end{figure}

\textbf{Manual Transcription and Spatial Annotation.} The retained samples are transcribed by 3 volunteer native Arabic speakers, who also generate bounding boxes using a free, open-source application\footnote{ \url{https://www.makesense.ai}} to capture the spatial alignment between words, characters, and stylistic forms. Since textual content varies across images, a single image may include multiple bounding boxes corresponding to distinct textual segments. Each sample is annotated with its transcription and style label, establishing a foundation for multimodal alignment between calligraphic form and semantic meaning as shown in Figure~\ref{fig:annotation_process}. To ensure annotation consistency, detailed guidelines were provided to annotators in advance (see Table~\ref{tab:annotation_guidelines} in Appendix~\ref{app:ver_anntate}).

\noindent\textbf{Normalization and De-duplication.} Then, samples undergo denoising and normalization to unify diacritics, letter variants like alef, hamza, ta'a marbouta, and encoding conventions to ensure that the corpus is both clean and standardized without compromising stylistic richness.

\subsection{Multi-tier Verification}
Quality assurance was conducted through a two-tier validation framework. Tier 1 addressed objective checks such as resolution, completeness, and artifact detection, while Tier 2 involved a calligraphy expert who verified transcription accuracy, diacritic integrity, and style and theme classification. After this process, the dataset was finalized with 1,272 high-quality samples. This layered approach combined systematic review with expert validation to ensure reliability and consistency of the annotations. More details on filtering criteria, verification steps, and annotation guidelines are provided in Appendix~\ref{app:ver_anntate}.\\

\vspace{-0.75em}

\subsection{Classification and Aggregation}
Finally, validated samples are classified by script type and thematic category before being aggregated into the final benchmark. Each sample is accompanied by structured metadata (including transcription, bounding box coordinates, word counts, and style annotations) forming the DuwatBench corpus. This process ensures that the dataset captures both visual diversity and semantic richness, making it a robust resource for multimodal learning. Figure~\ref{fig:samples_style} provides representative examples, illustrating how each image is paired with its corresponding annotations.

\begin{table*}[!ht]
\centering
\renewcommand{\arraystretch}{1.2}
\setlength{\tabcolsep}{10pt}
\resizebox{\textwidth}{!}{%
\begin{tabular}{llccccc}
\toprule
\rowcolor{gray!10} & \textbf{Models} & \textbf{CER\_mean} $\downarrow$ & \textbf{WER\_mean} $\downarrow$ & \textbf{chrF} $\uparrow$ & \textbf{ExactMatch} $\uparrow$  & \textbf{NLD\_error} $\downarrow$ \\

\midrule
\multirow{8}{*}{\rotatebox{90}{\small{Open-source}}} 
& Llava-v1.6-mistral-7b-hf \cite{liu2023improved} & 0.9932 & 0.9998 & 9.1582  & 0.0000 &  0.9114\\
& EasyOCR \cite{Kittinaradorn2024JaidedAI}     & 0.8538 & 0.9895 & 12.3016  & 0.0031 & 0.8163 \\
& InternVL3-8B \cite{chen2024internvl}  & 0.7588 & 0.8822 & 21.7461 & 0.0574 & 0.7132 \\
& Qwen2.5-VL-7B \cite{Qwen2.5-VL}               & 0.6453 & 0.7768 & 36.9703 & 0.1211 & 0.5984\\
& Qwen2.5-VL-72B-Instruct\cite{Qwen2.5-VL}  & 0.5709 & 0.7039 & \underline{43.9791} & 0.1761 & 0.5298\\
& Gemma-3-27B-IT \cite{team2025gemma}             & \textcolor{blue!70}{0.5556} & \textcolor{blue!70}{0.6591} & \textcolor{blue!70}{51.5330} & \textcolor{blue!70}{0.2398} & \textcolor{blue!70}{0.4741}\\
& trocr-base-arabic-handwritten\textsuperscript{\textcolor{red}{*}}
\cite{trocr_base_arabic_handwritten} & 0.9728 & 0.9998 & 1.7938  & 0.0000 & 0.9632\\
& MBZUAI/AIN\textsuperscript{\textcolor{red}{*} } \cite{heakl2025ainarabicinclusivelarge}                   & \underline{0.5494} & \underline{0.6912} & 42.6675 & \underline{0.1895} & \underline{0.5134} \\
\hline
\multirow{5}{*}{\rotatebox{90}{\small{Closed-source}}} 
& claude-sonnet-4.5 \cite{anthropic2025claude45}            & 0.6494 & 0.7255 & 42.9660 & 0.2225 & 0.5599\\
& gemini-1.5-flash \cite{gemini1.5}           & \underline{0.3933} & \underline{0.5112} & \underline{63.2790} & \underline{0.3522} & \underline{0.3659}\\
& gemini-2.5-flash \cite{gemini2.5flash}          & \textcolor{blue!70}{0.3700} & \textcolor{blue!70}{0.4478} &  \textcolor{blue!70}{71.8174} &  \textcolor{blue!70}{0.4167} &  \textcolor{blue!70}{0.3166}\\
& gpt-4o-mini \cite{gpt4omini}              & 0.6039 & 0.7077 & 42.6717 & 0.2115 & 0.5351 \\
& gpt-4o \cite{openai2024gpt4ocard}   & 0.4766 & 0.5692 & 56.8510 & 0.3388 & 0.4245\\
\bottomrule
\end{tabular}}
\vspace{-0.5 em}
\caption{
\textbf{Model Performance on DuwatBench.} Performance of open and closed-source models is reported using five complementary metrics: CER, WER, chrF, ExactMatch, and NLD. Results highlight clear differences across models, with larger multimodal models (gemini-2.5-flash and Gemma-3-27B-IT) showing stronger robustness, while others continue to face challenges on highly stylized calligraphic text.
\textit{Note:} \textcolor{red}{*} indicates Arabic-specific models. Values in blue denote \textcolor{blue!70}{the best} performance, while underlined values indicate \underline{the second-best} within each category (open or closed-source).}
\label{tab:main_results}
\vspace{-1em}
\end{table*}

\section{Benchmark Evaluation}

\subsection{Evaluation Metric}
We assess model performance with 5 complementary metrics: character error rate (CER), word error rate (WER), character F-score (chrF), ExactMatch, and normalized Levenshtein distance (NLD). CER captures the minimum number of character edits required to match the reference and is particularly important for Arabic calligraphy, where diacritics and ligatures can alter meaning. WER evaluates accuracy at the word level, offering a more interpretable measure for end-users. chrF computes character n-gram F-scores, rewarding partial overlaps and providing robustness to tokenization errors and spelling variants common in stylized text. ExactMatch is the strictest metric, counting only perfect matches between prediction and reference. Finally, NLD normalizes edit distance by sequence length, situating itself between CER and WER in granularity and offering a balanced perspective on recognition errors across variable word sizes.\\
Together, these metrics provide a balanced evaluation framework, capturing fine-grained character and word errors (CER, WER), normalized string similarity (NLD), partial correctness under visual distortion (chrF), and strict full-sequence accuracy (ExactMatch).

\vspace{-0.25em}
\subsection{Evaluation Settings}
All metrics were computed using standard open-source implementations to ensure reproducibility and consistency. The \textit{editdistance} and \textit{Levenshtein} libraries were used for computing character- and word-level edit operations (CER, WER, NLD), while \textit{sacrebleu} was employed for calculating chrF scores following machine translation evaluation standards. This setup provides a transparent and reproducible evaluation pipeline across all models.

To ensure fair comparison, we apply an Arabic-aware normalization pipeline before scoring. This includes Unicode normalization, removal of tatweel (\_), and unification of character variants such as Alef and hmaza forms and the Ya'a with Alef Maqsurah. Such preprocessing follows standard practice in Arabic NLP tools (e.g., CAMeL Tools, Lucene Arabic normalizer) and avoids penalizing models for superficial script variants.

\begin{table*}[!ht]
\centering
\renewcommand{\arraystretch}{1.2}
\setlength{\tabcolsep}{12pt}
\resizebox{\textwidth}{!}{%
\begin{tabular}{llcccccc}
\toprule
\rowcolor{gray!10}
& \textbf{Models} & \textbf{Kufic} & \textbf{Thuluth} & \textbf{Diwani} & \textbf{Naskh} & \textbf{Ruq'ah} & \textbf{Nasta'liq} \\
\midrule
\multirow{8}{*}{\rotatebox{90}{\small{Open-source}}}
& Llava-v1.6-mistral-7b-hf \cite{liu2023improved} 
& 1.0000 & 0.9996 & 1.0000 & 1.0000 & 1.0000 & 1.0000 \\
& EasyOCR \cite{Kittinaradorn2024JaidedAI}& 0.9895 & 0.9993 & 0.9990 & 0.9172 & 0.9757 & 0.9880  \\
& InternVL3-8B \cite{chen2024internvl}                            & 0.9416 & 0.8584 & 0.8537 & 0.9260 & 0.9973 & 0.9438  \\
& Qwen2.5-VL-7B \cite{Qwen2.5-VL}                                 & 0.8942 & 0.7578 & 0.8184 & 0.6443 & 0.8324 & 0.8583  \\
& Qwen2.5-VL-72B-Instruct \cite{Qwen2.5-VL}                       & 0.8658 &  \underline{0.6764} & 0.7640 & 0.5593 & 0.7613 & 0.7662\\
& Gemma-3-27B-IT \cite{team2025gemma}                             & \textcolor{blue!70}{0.7802} & \textcolor{blue!70}{0.6315} &  \underline{0.7326} & \textcolor{blue!70}{0.5138} &  \underline{0.7571} & \textcolor{blue!70}{0.6637} \\
& trocr-base-arabic-handwritten\textsuperscript{\textcolor{red}{*}} 
\cite{trocr_base_arabic_handwritten}     & 1.0000 & 0.9997 & 1.0000 & 0.9996 & 1.0000 & 1.0000 \\
& MBZUAI/AIN\textsuperscript{\textcolor{red}{*}} \cite{heakl2025ainarabicinclusivelarge} &  \underline{0.7916} & 0.7036 & \textcolor{blue!70}{0.7130} &  \underline{0.5367} & \textcolor{blue!70}{0.6111} &  \underline{0.6916} \\
\hline
\multirow{5}{*}{\rotatebox{90}{\small{Closed-source}}}
& claude-sonnet-4.5 \cite{anthropic2025claude45}                  & 0.8965 & 0.6724 & 0.8219 & 0.5859 & 0.8893 & 0.7659 \\
& gemini-1.5-flash \cite{gemini1.5}                                &  \underline{0.7212} &  \underline{0.4741} &  \underline{0.5783} &  \underline{0.4444} & \textcolor{blue!70}{0.5445} &  \underline{0.5023} \\
& gemini-2.5-flash \cite{gemini2.5flash}                           & \textcolor{blue!70}{0.7067} & \textcolor{blue!70}{0.3527} & \textcolor{blue!70}{0.5698} & 0.4765 & 0.5817 & 0.5222 \\
& gpt-4o-mini \cite{gpt4omini}                                     & 0.8790 & 0.6991 & 0.7825 & 0.5016 & 0.6908 & 0.6428 \\
& gpt-4o \cite{openai2024gpt4ocard}                                & 0.8041 & 0.5540 & 0.6370 & \textcolor{blue!70}{0.4189} &  \underline{0.5507} & \textcolor{blue!70}{0.4434} \\
\bottomrule
\end{tabular}}
\vspace{-0.5em}
\caption{
\textbf{WER Mean across Arabic calligraphy styles on DuwatBench.}  
WER scores are reported per script style (Kufic, Thuluth, Diwani, Naskh, Ruq'ah, Nasta'liq) for open and closed-source models.  
Lower values indicate better recognition accuracy.  
\small\textit{Note:} \textcolor{red}{*} indicates Arabic-specific models. Values in blue denote \textcolor{blue!70}{the best} performance, while underlined values indicate \underline{the second-best} within each category (open or closed-source).}
\label{tab:wer_style_results}
\vspace{-0.5em}
\end{table*}

\begin{table*}[!ht]
\centering
\renewcommand{\arraystretch}{1.2}
\setlength{\tabcolsep}{10pt}
\resizebox{\textwidth}{!}{%
\begin{tabular}{llccccc}
\toprule
\rowcolor{gray!10} & \textbf{Models} & \textbf{CER\_mean} $\downarrow$ & \textbf{WER\_mean} $\downarrow$ & \textbf{chrF} $\uparrow$ & \textbf{ExactMatch} $\uparrow$  & \textbf{NLD\_error} $\downarrow$ \\
\midrule
\multirow{8}{*}{\rotatebox{90}{\small{Open-source}}}
& Llava-v1.6-mistral-7b-hf \cite{liu2023improved} & 0.9998 & 1.0000 & 7.0348 & 0.0000 & 0.9420 \\
& EasyOCR \cite{Kittinaradorn2024JaidedAI}                & 0.8935 & 0.9917 & 12.0570 & 0.0018 & 0.8330 \\
& InternVL3-8B-Instruct \cite{chen2024internvl}           & 0.7330 & 0.8566 & 22.6430 & 0.0957 & 0.7021 \\
& Qwen2.5-VL-7B-Instruct \cite{Qwen2.5-VL}                & 0.5957 & 0.7341 & 37.8491 & 0.1805 & 0.5602 \\
& Qwen2.5-VL-72B-Instruct \cite{Qwen2.5-VL}               & 0.5683 & 0.6898 & 42.9576 & 0.2201 & 0.5242 \\
& Gemma-3-27B-IT \cite{team2025gemma}                     & \underline{0.5494} & \textcolor{blue!70}{0.6572} & \textcolor{blue!70}{48.9358} & \textcolor{blue!70}{0.2646} & \textcolor{blue!70}{0.4707} \\
& trocr-base-arabic-handwritten\textsuperscript{\textcolor{red}{*}}
\cite{trocr_base_arabic_handwritten}                      & 0.9746 & 0.9997 & 1.9925  & 0.0000 & 0.9626 \\
& MBZUAI/AIN\textsuperscript{\textcolor{red}{*}} \cite{heakl2025ainarabicinclusivelarge}
& \textcolor{blue!70}{0.5313} &  \underline{0.6692} &  \underline{43.9536} &  \underline{0.2233} &  \underline{0.4912} \\
\hline
\multirow{5}{*}{\rotatebox{90}{\small{Closed-source}}}
& claude-sonnet-4.5 \cite{anthropic2025claude45}          & 0.6896 & 0.7523 & 36.9575 & 0.2134 & 0.6076 \\
& gemini-1.5-flash \cite{gemini1.5}                        &  \underline{0.4110} &  \underline{0.5217} &  \underline{58.0771} &  \underline{0.3738} &  \underline{0.3823} \\
& gemini-2.5-flash \cite{gemini2.5flash}                   & \textcolor{blue!70}{0.3693} & \textcolor{blue!70}{0.4531} & \textcolor{blue!70}{66.5662} & \textcolor{blue!70}{0.4488} & \textcolor{blue!70}{0.3212} \\
& gpt-4o-mini \cite{gpt4omini}                              & 0.5975 & 0.6995 & 41.3582 & 0.2317 & 0.5290 \\
& gpt-4o \cite{openai2024gpt4ocard}                         & 0.6107 & 0.7010 & 40.6754 & 0.2463 & 0.5437 \\
\bottomrule
\end{tabular}}

\vspace{-0.5 em}
\caption{
\textbf{Model Performance on DuwatBench with Bounding Boxes.} The table reports CER, WER, chrF, ExactMatch, and NLD for open- and closed-source models evaluated on DuwatBench using bounding box annotations. gemini-2.5-flash maintains the best overall performance across metrics, while within open-source models, MBZUAI/AIN achieves the lowest CER, outperforming Gemma-3-27B-IT.
  \small\textit{Note:} \textcolor{red}{*} indicates Arabic-specific models. Values in blue denote \textcolor{blue!70}{the best} performance, while underlined values indicate \underline{the second-best} within each category (open or closed-source).
}
\label{tab:bbox_results}
\vspace{-1.15em}
\end{table*}

\begin{figure*}[t!]
\centering
\includegraphics[width=\textwidth,height=15cm]{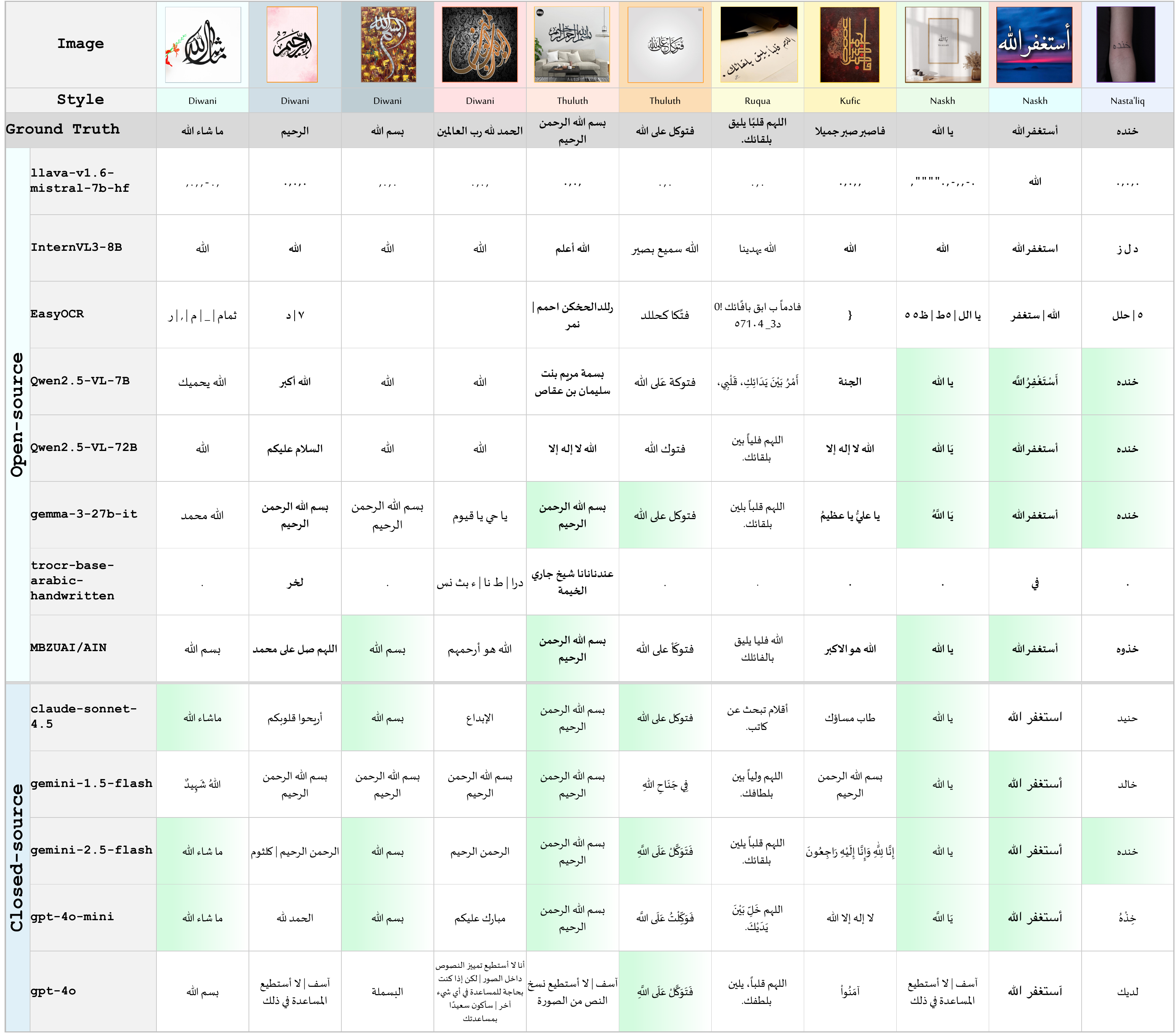}
\vspace{-1.5em}
\caption{
\small
\textbf{Qualitative Comparison of Model Outputs Across Calligraphic Styles}
Examples from DuwatBench showing transcription quality across six calligraphic styles, evaluated using 13 open and closed-source OCR and vision-language models. Each column presents the model’s predicted text aligned with the ground truth. The figure highlights challenges such as stroke density, curved baselines, and complex letter connections that complicate recognition, particularly in Thuluth, Diwani, and Kufic scripts. \emph{Note:} Cells highlighted in green indicate correct predictions where diacritic variations are ignored; inaccurate or misplaced hamza marks are considered errors.}
\label{fig:qual_grid}
\vspace{-1.25em}
\end{figure*}

\section{Results, Analysis, and Error Analysis}

\subsection{Quantitative Results}
Table~\ref{tab:main_results} presents the performance of open and closed-source models evaluated directly on full calligraphy images. The results reveal clear performance gaps, reflecting the models’ varying capacity to interpret the artistic and structural complexity of Arabic calligraphy.

Among open-source systems, Gemma-3-27B-it achieved the best overall performance, recording the lowest CER (0.56) and WER (0.66) and the highest chrF (51.53) and ExactMatch (0.24) within this group. Followed by MBZUAI/AIN and Qwen2.5-VL-72B-Instruct, with balanced results across all metrics, indicating their stronger adaptation to the Arabic script. In contrast, traditional OCR baselines such as EasyOCR and trocr-base-arabic-handwritten performed poorly, struggling with the ornate ligatures and curved baselines typical of calligraphy. Similarly, general-purpose LMMs such as LLaVA-v1.6-Mistral-7B underperformed, suggesting that instruction tuning without explicit Arabic grounding is insufficient for stylized text recognition.

Among closed-source systems, Gemini-2.5-Flash achieves the strongest overall performance, with the highest chrF (71.82), the best ExactMatch (0.42), and the lowest NLD error of (0.32), outperforming all baselines. Gemini-1.5-Flash ranks second, showing consistent robustness and surpassing GPT-4o and GPT-4o-mini in ExactMatch and NLD, while Claude-Sonnet-4.5 attains moderate performance.

\vspace{-0.5em}
Overall, these findings confirm that large, instruction-optimized multimodal models, particularly Gemini-2.5-Flash and Gemma-3-27B-IT, are more resilient to the structural irregularities and stylistic diversity of Arabic calligraphy than conventional OCR tools or smaller open models. This highlights the importance of large-scale multilingual pretraining and cross-modal grounding for robust Arabic visual text understanding.

Table~\ref{tab:wer_style_results} further reports WER across the six DuwatBench calligraphic styles. Models perform best on Naskh and Ruq'ah, particularly among open-source systems, where strokes are standardized and geometric distortion is limited. In contrast, ornate scripts such as Diwani and Thuluth remain challenging due to dense ligatures and fluid baselines. Nasta'liq exhibits variances, with some models achieving low error while others fail markedly, suggesting uneven exposure to this script during pretraining. Among open-source models, Gemma-3-27B-IT shows the most consistent cross-style performance, while MBZUAI/AIN records the lowest errors for Diwani and Ruq'ah, indicating familiarity with the two styles and modern Arabic handwriting. Within closed-source systems, Gemini-2.5-Flash, Gemini-1.5-Flash, and GPT-4o exhibit relatively stable performance across styles. Nevertheless, occasional outliers, particularly on Naskh, Ruq'ah, and Nasta'liq, suggest variability in script-specific robustness, likely reflecting differences in script coverage during pretraining (additional results in Table~\ref{tab:chrf_style_results}, Appendix~\ref{app:eval_more}).

\vspace{-0.5em}
These results reveal that model accuracy correlates strongly with style prevalence in pretraining data. Scripts such as Naskh and Ruq'ah, which dominate modern Arabic documents, yield the highest recognition rates, whereas historical and ornamental styles such as Kufic and Diwani remain the most error-prone.

To examine the impact of localization, we conducted ablation experiments in which models were evaluated using bounding box guidance that isolates calligraphic text regions. As shown in Table~\ref{tab:bbox_results}, this setting improves recognition accuracy for the majority of models, with particularly notable gains in ExactMatch and NLD, indicating more accurate holistic transcription when background clutter is reduced. Several open-source models, including Gemma-3-27B-IT, Qwen2.5-VL-72B-Instruct, and MBZUAI/AIN, show consistent improvements under localized input, suggesting that focused visual grounding better supports fine-grained character recognition in stylized scripts. Among closed-source systems, Gemini-2.5-Flash remains the strongest performer in both settings, maintaining robust accuracy with and without bounding boxes. In contrast, traditional OCR-based methods (EasyOCR, TrOCR) remain weak, underscoring their limited capacity to generalize beyond printed or lightly cursive Arabic text.
While bounding box cropping mitigates background interference, cross-style imbalance persists, emphasizing the need for targeted data augmentation and style-aware pretraining.

In general, incorporating bounding box localization leads to consistent performance gains across most models, particularly enhancing recognition accuracy and alignment in visually complex calligraphic text. Nevertheless, fully addressing the challenges posed by Arabic calligraphy requires continued efforts toward improving coverage of underrepresented styles and reducing pretraining bias.

\subsection{Error Analysis}
\textbf{Quantitative Error Analysis.}
Table~\ref{tab:quantitative_error_analysis} summarizes the performance of the models in the five metrics. The correlation analysis (see Figure~\ref{fig:metric_correlation} in Appendix~\ref{app:error_analysis}) reveals strong relationships between character and word errors ($\rho_{CER,WER}=0.98$), as well as strong inverse correlations between normalized edit distance and semantic overlap (NLD–chrF = –0.99, NLD–ExactMatch = –0.95), indicating that lexical deviations closely align with losses in visual and semantic fidelity.

Closed-source models consistently outperform open-source counterparts, particularly in normalized edit distance ($p=0.0112$) and exact-match accuracy ($p=0.0029$); however, all systems struggle with complex scripts and long inscriptions. Performance degrades as word count and stylistic curvature increase, with the largest errors observed in Thuluth and Diwani samples, where ligatures, flourishes, and overlapping baselines amplify recognition difficulty. In contrast, models perform more reliably on shorter phrases and on Naskh and Ruq'ah texts, which exhibit more linear structure and consistent spacing.

Across the dataset, average ExactMatch remains below 0.18, underscoring the persistent challenge of faithfully transcribing stylized Arabic calligraphy even for advanced multimodal LLMs. These quantitative trends align with the qualitative analysis (Figure~\ref{fig:qual_grid}), where increased stylistic and visual complexity directly corresponds to recognition failures. Together, these results validate the proposed metric suite and provide a rigorous quantitative basis for cross-model performance comparison on DuwatBench. More details are provided in Appendix~\ref{app:error_analysis}.

\textbf{Qualitative Error Analysis.}
Figure~\ref{fig:qual_grid} visualizes representative outputs across six calligraphic styles, highlighting correct predictions in green. The results show substantial variation between models and styles. Among open-source systems, LLaVA-v1.6-Mistral-7B often produces incoherent Arabic, InternVL3-8B tends to overgenerate generic tokens like "Allah", and EasyOCR exhibits fragmented outputs. Gemma-3-27B-IT achieves the most stable open-source performance, while Qwen2.5-VL variants perform best on simpler scripts like Naskh and Ruq'ah. MBZUAI/AIN performs competitively for its smaller scale, notably producing a correct Diwani recognition case, an area where most other open-source models fail.

In the closed-source category, Claude-Sonnet-4.5 and Gemini-2.5-Flash produce the most reliable transcriptions across Diwani and Thuluth, while GPT-4o-mini remains consistent, albeit slightly weaker, on Thuluth. GPT-4o frequently rejects input, reflecting stricter safety filtering.

Certain qualitative patterns also emerge. Models occasionally generate semantically plausible but visually mismatched outputs. For example, Gemma-3-27B-IT correctly recognized  "Bismillah" but expanded it to the full Basmala  "Bismillah Ar Rahman Ar Rahim" indicating cultural priors influencing prediction. Across models, a tendency to overpredict  "Allah" reflects strong cultural associations learned during training. Meanwhile, styles like Kufic and Diwani, which are geometrically rigid or ornate, remain the hardest to decode, while Naskh, Ruq'ah, and Nasta'liq achieve the highest recognition stability.

In summary, these findings highlight both linguistic and cultural biases in Arabic multimodal models and show that visual-textual grounding (rather than model size alone) is key to advancing calligraphy understanding.

\section{Conclusion}
In this work, we introduce DuwatBench, a benchmark for Arabic calligraphy comprising more than 1.27K curated samples that span various text categories and calligraphic styles. The dataset combines full transcriptions, style annotations, and detection-level information with authentic artistic backgrounds, offering a realistic testbed for Arabic LMMs. Our evaluation shows that, while existing systems handle standard text, they struggle with the stylistic variation and visual complexity of calligraphy. The benchmark highlights these limitations and encourages future research on script-aware modeling and culturally grounded AI, with applications in cultural preservation, education, and digital humanities.

\section{Limitations and Societal Impact}
Our proposed benchmark is designed as a focused and high-quality resource for evaluating Arabic calligraphy in multimodal models. Although the scale of the data set is smaller than that of large general-purpose Arabic corpora, it emphasizes diversity between text categories and calligraphic styles and ensures meaningful evaluation scenarios. Beyond technical contributions, the dataset highlights the cultural and artistic significance of Arabic calligraphy and supports research in areas such as cultural heritage preservation, digital archiving, and education. With this release, we aim to promote responsible and inclusive research practices that respect the cultural significance of the Arabic script while advancing the development of robust multimodal systems.

\bibliography{arxiv}

\clearpage
\appendix

\section{Appendix}
\label{sec:appendix}
This appendix provides supplementary material for our study of Arabic calligraphy understanding in multimodal models. It includes: (1) an overview of related work situating DuwatBench within existing resources; (2) detailed dataset statistics; (3) verification and validation guidelines used by annotators and experts; (4) additional evaluation results; and (5) supplementary statistical analysis highlighting the dataset’s stylistic and cultural diversity. Together, these materials emphasize DuwatBench’s role in bridging linguistic meaning with visual heritage and providing a solid foundation for model evaluation.

\section{Related Work}
\label{app:rw}
Research on Arabic calligraphy intersects with Natural Language Processing (NLP), multimodal learning, and cultural heritage. Although prior work has produced valuable benchmarks and datasets, most focus on clean text, structured documents, or narrow stylistic slices, leaving the artistic and cultural complexity of calligraphy largely unaddressed. We group related work into three areas: (i) Arabic multimodal benchmarks, (ii) heritage and cultural resources, and (iii) Arabic calligraphy datasets.

\subsection{Arabic Multimodal Benchmarks}

Arabic multimodal benchmarks have evolved from general-purpose evaluation suites to more specialized resources. CAMEL-Bench established a large-scale benchmark covering eight domains with over 29K native-curated questions~\cite{ghaboura2024camel}. Dialectal variation was later emphasized through resources such as Dallah~\cite{alwajih2024dallah} and JEEM~\cite{kadaoui2025jeemvisionlanguageunderstandingarabic}, underscoring the importance of capturing linguistic diversity across the Arab world. Building on these directions, the ARB benchmark~\cite{ghaboura2025arb} introduced step-by-step multimodal reasoning across 11 domains, spanning visual, OCR, and document understanding challenges. Although these resources significantly advance Arabic multimodal research, they remain centered on language, dialect, and document-level OCR. None address the stylistic and artistic dimensions of Arabic script, where the written form functions as both text and art.
\vspace{-0.5em}
\subsection{Heritage and Cultural Benchmarks}

Heritage-oriented benchmarks demonstrate how evaluation can extend beyond functional NLP to cultural knowledge and artistic expression. For example, Fann or Flop~\cite{alghallabi2025fann} foregrounds Arabic poetry, requiring models to interpret figurative language and aesthetic conventions across eras and genres. Other initiatives, such as Sacred or Synthetic ~\cite{atif2025sacred}, PalmX~\cite{alwajih2025palmx}, and PEARL~\cite{alwajih2025pearl}, introduced shared tasks for benchmarking LLMs on Arabic and Islamic cultural knowledge, underscoring the importance of cultural grounding. These resources have advanced heritage-aware evaluation, but they largely address poetry, artifacts, or cultural reasoning, while the artistic and stylistic dimension of Arabic calligraphy remains overlooked.

\subsection{Arabic Calligraphy Resources}

Computational work on Arabic calligraphy has primarily targeted recognition, identification, and style classification. Early studies employed handcrafted features to distinguish between scripts~\cite{kaoudja2020arabic,ezz2019classification}, but their impact was limited by small or restricted-use datasets. Later contributions introduced larger collections to support OCR and style recognition at the word and manuscript level~\cite{alyafeai2022calliar,ismail2023hicma,adam2017based,zoizou2022moj,saeed2024muharaf}, complemented by computational approaches for style representation~\cite{kaoudja2021new}. Earlier computer vision methods also explored the recognition of artistic scripts~\cite{allaf2016automatic,salamah2018towards}, while recent surveys highlight interest in script generation through deep learning and generative models~\cite{sumaylihandwritten}. More recently, multimodal efforts such as Qalam~\cite{bhatia2024qalam} have signaled a transition from traditional OCR to vision–language approaches for Arabic handwriting.

Despite these contributions, prior datasets remain limited: they typically frame calligraphy as OCR rather than detection, rely on cropped text instead of full artistic layouts, focus mainly on historical manuscripts, and lack semantic diversity. \textbf{DuwatBench} addresses these gaps by introducing 1,272 samples across six major styles with bounding box annotations and a wide range of textual categories. Unlike earlier corpora, its samples appear in complex artistic contexts such as inscriptions, signage, and artworks, foregrounding the visual–linguistic complexity of calligraphy in real-world settings.

\section{Data Statistics}
\label{app:data_stat}
\begin{figure}[!ht]
\centering
\includegraphics[width=0.8\linewidth]{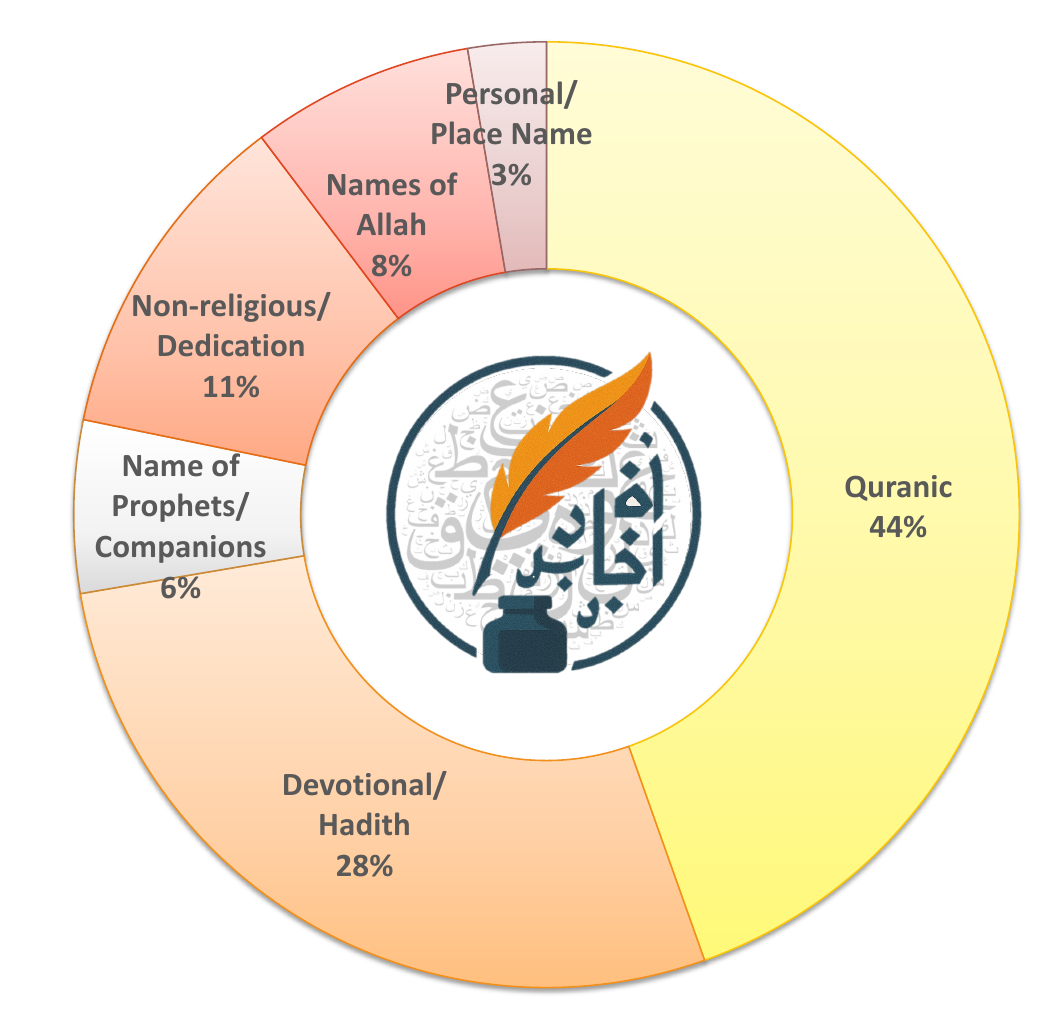}
\vspace{-0.75em}
\caption{
\small
\textbf{DuwatBench Theme-level Distribution.} Distribution of DuwatBench textual content across thematic categories, including non-religious text, Quranic verses, devotional invocations, names of the Prophet and companions, Names of Allah, and smaller groups such as person names or places, dedications, and hadith.}
\vspace{-0.75em}
\label{fig:theme_stat}
\end{figure}

\begin{figure}[!ht]
\centering
\includegraphics[width=0.8\linewidth]{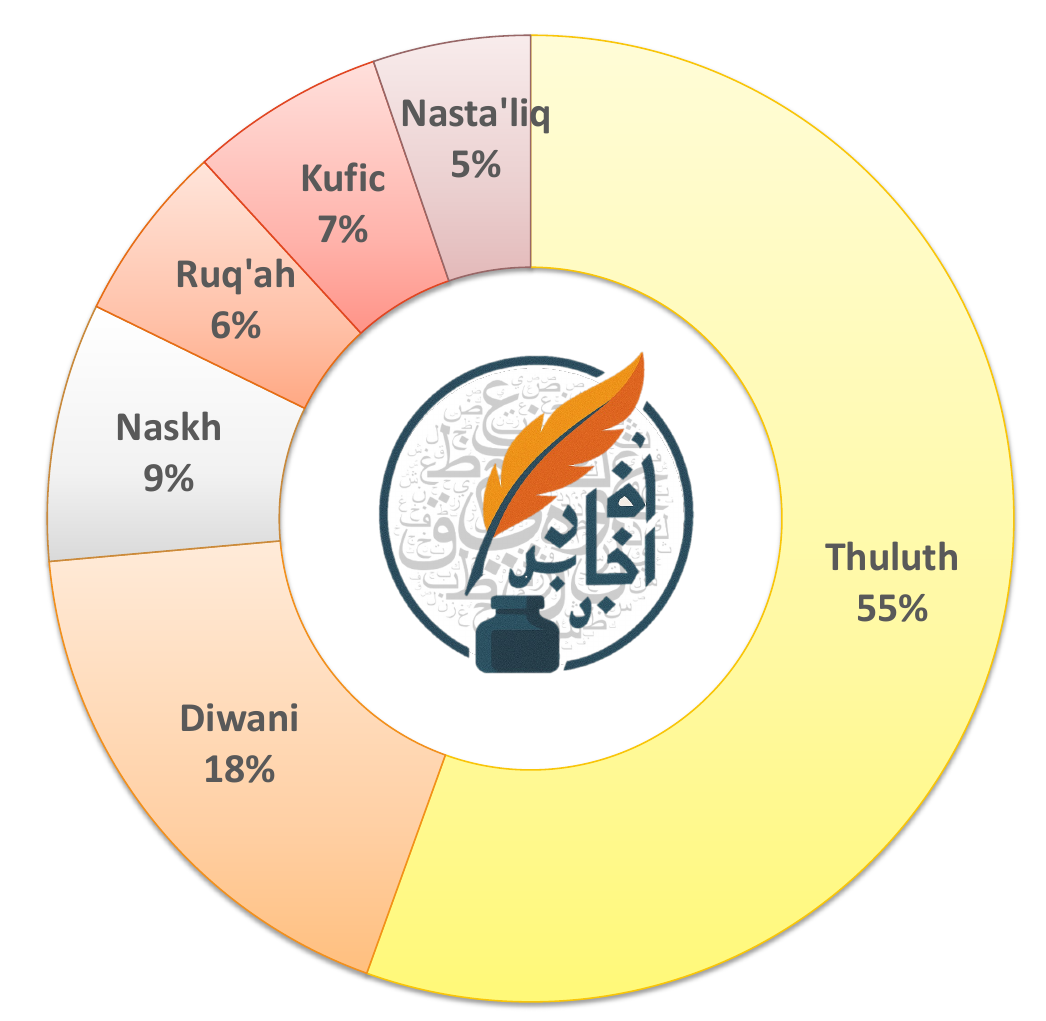}
\vspace{-0.75em}
\caption{
\small
\textbf{DuwatBench Style-level Distribution.} Distribution of DuwatBench samples across six major Arabic calligraphy styles, with Thuluth and Diwani forming the majority and Kufic, Naskh, Ruq'ah, and Nasta'liq represented in smaller proportions.}
\vspace{-0.5em}
\label{fig:style_stat}
\end{figure}

To better characterize the distributional properties of our Arabic calligraphy dataset, we report descriptive statistics across both stylistic and thematic dimensions. Figures~\ref{fig:theme_stat} and~\ref{fig:style_stat} illustrate that the corpus is primarily composed of widely used scripts such as Thuluth and Diwani, with comparatively lower representation of Kufic, Naskh, Ruq'ah, and Nasta'liq.

 On the content side, the dataset includes both non-religious expressions and religious material, with a strong emphasis on Quranic verses, devotional phrases, divine names, and prophetic references. This distribution underscores the dual role of Arabic calligraphy as both a cultural and sacred medium, while capturing the diversity needed for robust multimodal evaluation.
\vspace{-0.5em}
\begin{figure}[!ht]
\centering
\includegraphics[width=\linewidth]{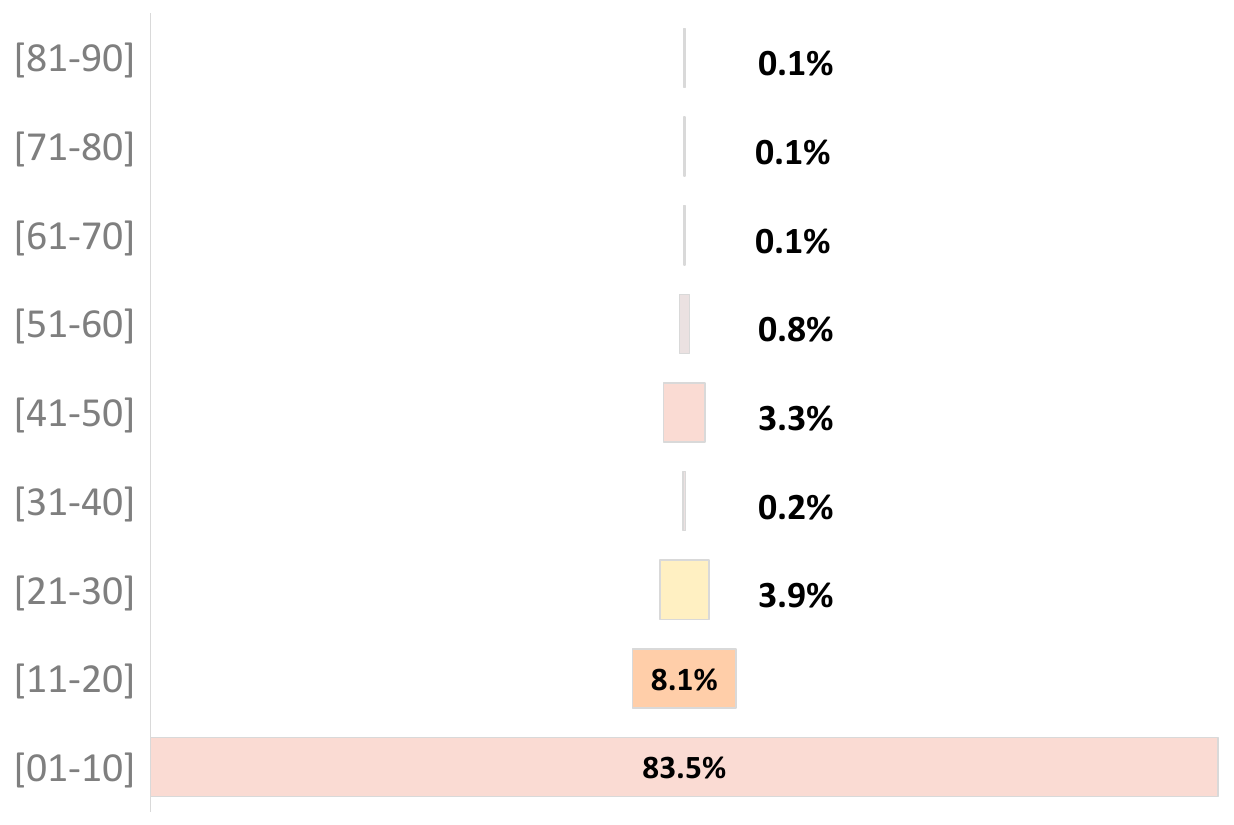}
\vspace{-2em}
\caption{
\small
\textbf{Distribution of total word counts per entry in DuwatBench.} While most samples in DuwatBench consist of single words (83.5\%), the dataset also includes longer inscriptions, particularly Quranic verses and combined passages.}
\vspace{-0.5em}
\label{fig:word_count}
\end{figure}

Figure~\ref{fig:word_count} illustrates the distribution of word counts across entries in DuwatBench. The majority of samples are single words, reflecting the common use of calligraphy for standalone terms such as names, motifs, or short expressions. At the same time, the dataset captures longer and more complex inscriptions, especially in religious contexts. Examples include extended Quranic verses such as Ayat al-Kursi and Al-Fatihah, as well as composite renderings that combine multiple surahs—for instance, An-Nas, Al-Falaq, and Al-Ikhlas presented alongside Ayat al-Kursi. This range highlights the dataset’s ability to represent both concise and extended forms of calligraphy, ensuring coverage of diverse inscription practices.

\section{Verification and Annotation Guidelines}
\label{app:ver_anntate}
To ensure consistency and reliability across annotations, we designed a clear set of guidelines for annotators and validators. These rules covered image inclusion, box creation, transcription practices, and style/theme labeling, followed by a two-tier verification process. Table~\ref{tab:annotation_guidelines} summarizes the rubric applied during the annotation and verification phase of DuwatBench.

\begin{table*}[!t]
\centering
\small 
\renewcommand{\arraystretch}{1.2}
\setlength{\tabcolsep}{6pt}
\begin{tabular}{p{0.28\textwidth} p{0.65\textwidth}}
\toprule
\rowcolor{gray!10} \textbf{Aspect} & \textbf{Guidelines} \\
\midrule
\textbf{Image inclusion} &
- Good resolution. \newline
- Calligraphy must be complete, legible, and undistorted. \newline
- Reject blurred, overexposed, cropped, or incomplete text. \newline
- Prefer authentic artistic backgrounds preserving natural context. \\
\midrule
\textbf{Bounding boxes} &
- Draw boxes tightly around each word. \newline
- Do not cut diacritics or ornamental strokes. \newline
- Avoid excessive empty space around text. \\
\midrule
\textbf{Transcription} &
- Manually transcribe all visible text. \newline
- Normalize letter forms and diacritics for consistency. \newline
- Use Unicode-compliant Arabic script. \\
\midrule
\textbf{Style \& theme labeling} &
- Assign one of six calligraphic styles: Thuluth, Diwani, Kufic, Naskh, Ruq'ah, Nasta'liq. \newline
- Assign a thematic category: non-religious, Quranic, devotional, Names of Allah, prophetic references, or smaller groups (personal dedications, hadith). \\
\midrule
\textbf{Quality control (Tier 1)} &
- Check resolution, completeness of text, and absence of artifacts. \\
\midrule
\textbf{Expert validation (Tier 2)} &
- Verify transcription accuracy and diacritic placement. \newline
- Confirm style and theme classification. \newline
- Resolve disagreements through consensus. \\
\bottomrule
\end{tabular}
\vspace{-0.5em}
\caption{\textbf{DuwatBench Verification and Annotation Guidelines.} The table summarizes the procedures followed during dataset verification and labeling. It outlines key aspects of the annotation process from image inclusion criteria and bounding box placement to transcription, style, and thematic labeling, alongside a two-tier validation framework. Tier~1 ensures visual and structural quality control, while Tier~2 involves expert validation of transcription accuracy, diacritic integrity, and style–theme consistency.}
\label{tab:annotation_guidelines}
\end{table*}

\section{Additional Results and Analysis}
\label{app:eval_more}

\begin{table*}[!ht]
\centering
\renewcommand{\arraystretch}{1.2}
\setlength{\tabcolsep}{12pt}
\resizebox{\textwidth}{!}{%
\begin{tabular}{llcccccc}
\toprule
\rowcolor{gray!10}
& \textbf{Models} & \textbf{Kufic} & \textbf{Thuluth} & \textbf{Diwani} & \textbf{Naskh} & \textbf{Ruq'ah} & \textbf{Nasta'liq}\\
\midrule
\multirow{8}{*}{\rotatebox{90}{\small{Open-source}}}
& Llava-v1.6-mistral-7b-hf \cite{liu2023improved}& 7.1477 & 10.0517 & 7.1070 & 8.5065 & 7.7523 & 10.4573  \\
& EasyOCR \cite{Kittinaradorn2024JaidedAI}& 11.0156 & 9.5411 & 5.5714 & 37.5340 & 19.9125 & 16.0816 \\
& InternVL3-8B \cite{chen2024internvl}& 15.5629 & 23.4338 & 24.2987 & 21.4362 & 11.6528 & 15.9995 \\
& Qwen2.5-VL-7B \cite{Qwen2.5-VL}& 21.5025 & 37.6000 & 30.4223 & 57.2947 & 37.8429 & 37.4302\\
& Qwen2.5-VL-72B-Instruct \cite{Qwen2.5-VL}& 24.0088 &  \underline{45.4786} & 35.1459 &  \underline{65.2360} &  \underline{49.1668} &  \underline{42.4100} \\
& Gemma-3-27B-IT \cite{team2025gemma}& \textcolor{blue!70}{37.1295} & \textcolor{blue!70}{52.6964} & \textcolor{blue!70}{45.8464} & \textcolor{blue!70}{69.3757} & 47.0942 & \textcolor{blue!70}{53.8704} \\
& trocr-base-arabic-handwritten\textsuperscript{\textcolor{red}{*}} 
\cite{trocr_base_arabic_handwritten}& 1.2777 & 1.5063 & 2.5502 & 1.2485 & 3.3519 & 2.3474 \\
& MBZUAI/AIN\textsuperscript{\textcolor{red}{*}} \cite{heakl2025ainarabicinclusivelarge} &  \underline{31.7205} & 40.7797 &  \underline{40.0978} & 61.6722 & \textcolor{blue!70}{55.6625} & 42.1033 \\
\hline
\multirow{5}{*}{\rotatebox{90}{\small{Closed-source}}}
& claude-sonnet-4.5 \cite{anthropic2025claude45}    & 23.6520 & 48.6203 & 30.4111 & 57.0519 & 30.8645 & 40.8687 \\
& gemini-1.5-flash \cite{gemini1.5}  &  \underline{42.9874} &  \underline{63.3661} &  \underline{57.5302} & \textcolor{blue!70}{80.5358} & \textcolor{blue!70}{71.9229} & 68.9132  \\
& gemini-2.5-flash \cite{gemini2.5flash}  &  \textcolor{blue!70}{47.4052} & \textcolor{blue!70}{77.0677} &  \textcolor{blue!70}{62.1220} & 77.8096 &  \underline{69.2090} & \textcolor{blue!70}{71.2734}  \\
& gpt-4o-mini \cite{gpt4omini}& 27.9043 & 40.8158 & 35.0949 &  71.6135 &  52.1796 & 51.8428 \\
& gpt-4o \cite{openai2024gpt4ocard} & 35.0349 & 55.8203 & 50.5606 &  \underline{80.2030} & 67.5542 &   \underline{69.8079}\\
\bottomrule
\end{tabular}}
\vspace{-0.5em}
\caption{
\textbf{ChrF scores across Arabic calligraphy styles on DuwatBench.}  
ChrF values are reported per script style (Kufic, Thuluth, Diwani, Naskh, Ruq'ah, Nasta'liq) for open and closed-source models.  
Higher scores indicate better recognition quality.  
\small\textit{Note:} \textcolor{red}{*} indicates Arabic-specific models. Values in blue denote \textcolor{blue!70}{the best} performance, while underlined values indicate \underline{the second-best} within each category (open or closed-source).}
\label{tab:chrf_style_results}
\end{table*}

To complement the main evaluation, this section presents additional quantitative results across Arabic calligraphic styles using the chrF metric, offering a finer view of character-level recognition and how performance varies with script complexity and data representation.

Table~\ref{tab:chrf_style_results} shows notable variation in recognition difficulty. Kufic records the lowest scores overall, reflecting its geometric rigidity and sparse representation in training data. Thuluth and Diwani achieve moderate results—lower for open-source models but substantially higher for closed-source ones such as Gemini-2.5-Flash and Gemini-1.5-Flash. Naskh and Ruq'ah remain the most accurately recognized, while Nasta'liq yields moderate but inconsistent outcomes, performing better in multilingual models like Gemma-3-27B-IT, Gemini-2.5-Flash, and GPT-4o. Notably, MBZUAI/AIN demonstrates stable performance across scripts, leading in Ruq'ah and achieving competitive results in Diwani and Kufic, reflecting effective script-aware adaptation despite its comparatively smaller model size. These observations indicate that style complexity and data diversity, rather than model scale alone, play a decisive role in determining recognition quality for Arabic calligraphy.

\begin{table*}[!ht]
\centering
\small
\renewcommand{\arraystretch}{1.15}
\setlength{\tabcolsep}{4pt}
\begin{tabular}{lccccc}
\toprule
\rowcolor{gray!10}
\textbf{Metric} & \textbf{Mean} & \textbf{Std} & \textbf{Min} & \textbf{Max} & \textbf{Range / Rel. Impr.} \\
\midrule
CER\_mean & 0.6456 & 0.1993 & 0.3700 & 0.9932 & 0.6232 (62.75\%) \\
WER\_mean & 0.7434 & 0.1819 & 0.4478 & 0.9998 & 0.5520 (55.21\%) \\
NLD\_error & 0.5940 & 0.2017 & 0.3166 & 0.9632 & 0.6466 (67.13\%) \\
chrF & 38.2873 & 21.3941 & 1.7938 & 71.8174 & 70.0236 \\
ExactMatch & 0.1791 & 0.1390 & 0.0000 & 0.4167 & 0.4167 (41.7\%) \\
\midrule
\rowcolor{gray!10}
\textbf{t-test (Open vs Closed)} & CER\_p=0.0181 & WER\_p=0.0087 & NLD\_p=0.0112 & chrF\_p=0.0084 & ExactMatch\_p=0.0029 \\
\bottomrule
\end{tabular}
\vspace{-0.4em}
\caption{\textbf{Cross-Model Error Statistics and Significance Analysis.} 
Descriptive statistics and t-test comparisons between open and closed-source models on DuwatBench. Results show that closed-source models achieve significantly lower normalized edit distance and higher ExactMatch accuracy, indicating greater robustness to stylistic variation in Arabic calligraphy.}
\vspace{-0.5em}
\label{tab:quantitative_error_analysis}
\end{table*}

\section{Supplementary Statistical Analysis}
\label{app:error_analysis}

This section provides additional quantitative analysis of model performance on DuwatBench. It complements the main results by presenting descriptive statistics, relative improvements, and statistical comparisons between open and closed-source systems. Together, these analyses provide a deeper understanding of how different models handle stylized Arabic calligraphy across recognition metrics.

\begin{figure}[!h]
\centering
\includegraphics[width=0.48\textwidth,height=5.25cm]{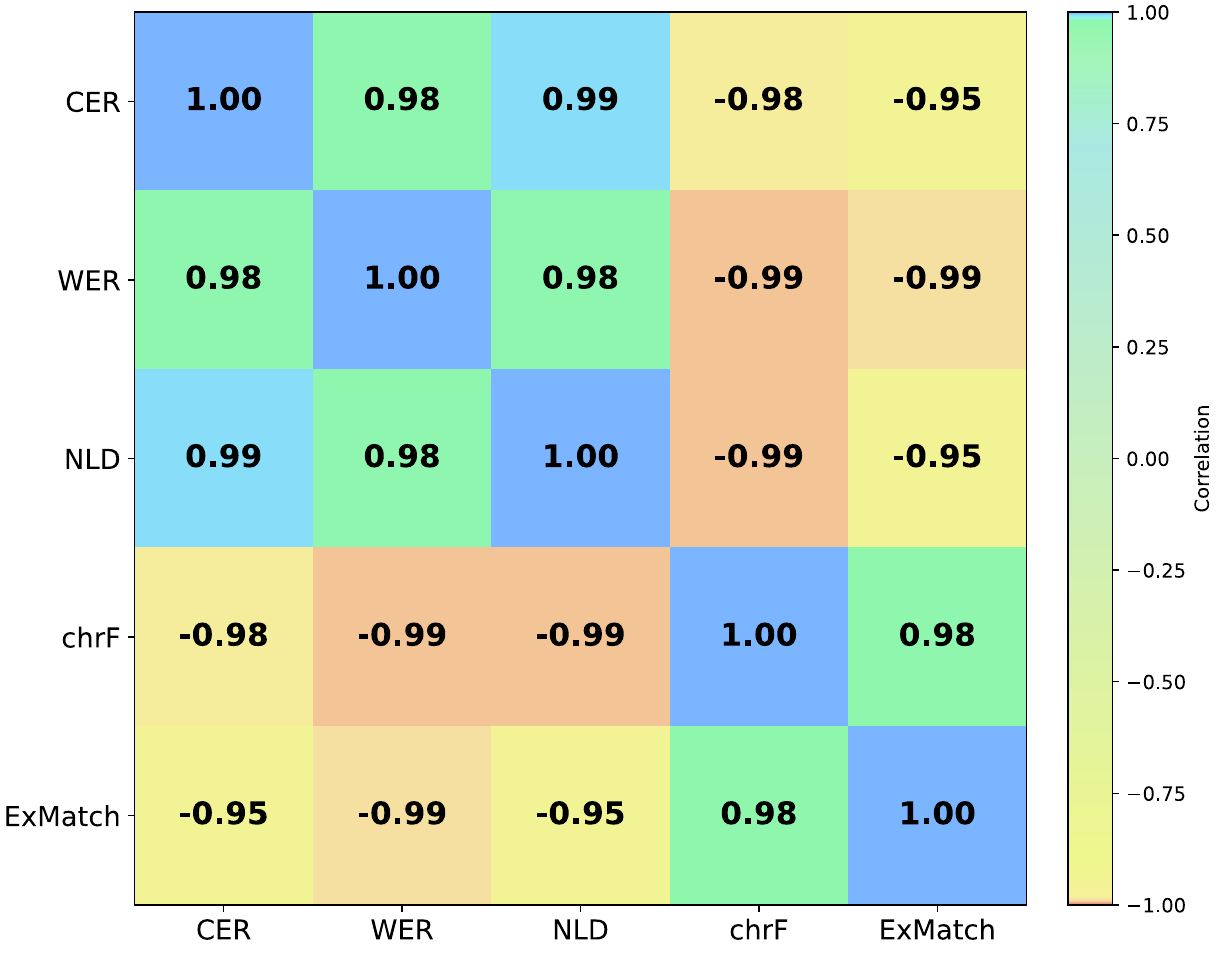}
\vspace{-1.25em}
\caption{
\small
\textbf{Metric Correlation Heatmap.} Correlation matrix across five metrics on DuwatBench, showing that higher edit distances (CER, WER, NLD) strongly inversely correlate with chrF and ExactMatch, confirming their complementary behavior in assessing recognition quality.
}
\label{fig:metric_correlation}
\vspace{-0.5em}
\end{figure}

Across all evaluated models, the average CER (0.65), and average WER (0.74) reflect the inherent difficulty of recognizing visually complex Arabic scripts. The mean NLD (0.59) indicates frequent partial mismatches, while mean chrF (38.29) and ExactMatch (0.18) suggest limited fluency and full-sequence accuracy. Large standard deviations ($\approx$ 0.18–0.2 for CER/WER/NLD and $\approx$ 21 for chrF) reveal substantial variability across systems, highlighting uneven robustness under diverse styles and artistic layouts.

Relative improvement (Rel. Impr.) between the strongest and weakest systems reaches 55–67\% for CER, WER, and NLD, underscoring the impact of model scale and multimodal grounding. Independent two-sample \textit{t}-test comparing open- and closed-source models (Table~\ref{tab:quantitative_error_analysis}) show that closed-source systems achieve significantly lower error rates across CER, WER, and NLD ($p<0.05$), as well as significantly higher chrF and ExactMatch scores ($p<0.01$), indicating consistent advantages at both character- and sequence-level evaluation, with the strongest gains in normalized edit distance and exact transcription accuracy.

These findings align with the qualitative analyses presented in the main paper, where closed-source models demonstrated greater stability on ornate scripts (e.g., Diwani, Thuluth), whereas open-source models exhibited higher variance and frequent segmentation errors. Together, Figure \ref{fig:metric_correlation} and Table \ref{tab:quantitative_error_analysis} provide a comprehensive quantitative view of model behavior, reinforcing the need for style-aware and culturally grounded evaluation of Arabic calligraphy recognition.

\end{document}